\documentclass[letterpaper]{article}
\usepackage{aaai}
\usepackage{times}
\usepackage{helvet}
\usepackage{courier}
\usepackage{url}
\usepackage{booktabs}
\usepackage{xspace}
\usepackage[pdftex]{graphicx}
\usepackage{subfigure,wrapfig}
\usepackage{enumitem}
\graphicspath{{images/}{images/proc_images/}}

\newenvironment{my_enumerate}{
\begin{enumerate}[leftmargin=*]
  \setlength{\itemsep}{1pt}
  \setlength{\parskip}{0pt}
  \setlength{\parsep}{0pt}}{\end{enumerate}
}

\newcommand{\ie}{\emph{i.e.}\xspace}
\newcommand{\eg}{\emph{e.g.}\xspace}
\newcommand{\etc}{\emph{etc.}\xspace}
\newcommand{\etal}{\emph{et al.}\xspace}

\begin{document}

\title{A Concept Learning Approach for Multisensory Object Perception}

\maketitle
\begin{abstract}
\begin{quote}
This paper presents a computational model of concept learning using Bayesian inference for a grammatically structured hypothesis space, and test the model on multisensory (visual and haptics) recognition of 3D objects. The study is performed on a set of artificially generated 3D objects known as fribbles, which are complex, multipart objects with categorical structures. The goal of this work is to develop a working multisensory representational model that integrates major themes on concepts and concepts learning from the cognitive science literature. The model combines the representational power of a probabilistic generative grammar with the inferential power of Bayesian induction.
\end{quote}
\end{abstract}

\section{Introduction}
In this paper, we propose a concept learning formulation for reasoning about complex 3-D objects, using multisensory perception information, specifically visual and haptic sensory data. We first explore the definitions of \emph{concepts}, \emph{rational rules}, and the \emph{language of thought} from the literature, and describe how these can form a basis for concept learning. We then extend these definitions to motivate and formulate our proposed computational approach for reasoning about objects using a multisensory perception system.

\subsection{Concepts, rational rules and the language of thought (LoT)}\label{sec:concepts}
In \cite{goodmanGFT07}, Goodman \etal presented  three standard intuitions relating to the notion of concepts, which we restate in order to motivate our computational approach to concept learning  for multisensory object perception.
\begin{my_enumerate}
\item ``Concepts are mental representations that are used to discriminate between objects, events, relations, or other states of affairs.'' Specifically in machine perception, we are concerned with those concepts that can be used to specify classes or categories of things.
\item ``Concepts are learned inductively from the sparse and noisy data of an uncertain world.... Even very sparse and noisy evidence, such as a few randomly encountered examples, can be sufficient for a young child to accurately grasp a new concept.'' This notion of learning about concepts from sparse noisy data is somewhat contrary to the philosophy embraced by many current state-of-the art approaches in machine perception, where the norm is to collect a very large number of examples for training an agent \cite{ikea13}. In this work, we describe a method of learning about object-related concepts from very few randomly selected instances, in a multisensory setting.
\item ``Many concepts are formed by combining simpler concepts, and the meanings of complex concepts are derived  in systematic ways from the meanings of their constituents.'' This intuition about concepts attempts to explain how a rational agent agent (human or machine) is able to elicit complex meanings from the raw sensory data that it perceives. In this work, we first separate the raw sensory data perceived from the governing concepts and then unify them in the proposed computational framework.
\end{my_enumerate}
One early observation to make about concepts thus far, is the need for a set of rules to define that concept. We refer to these as the \emph{rational rules}, the set of rules for designing and regulating a system, based on technical knowledge, and with the aim of achieving optimum outcomes. Rational rules can be based on universal laws such as the laws of physics, laws of morality, \etc, or on more domain specific principles such as Gestalt principles of visual perception, but whatever the domain-of-interest, these rational rules serve to provide an agent with a belief system or a framework to make predictions about how the world works. Thus, when given a statement about the world, an agent can determine its degree of truthfulness based on this belief system. For clarity, we define an agent\footnote{By believing that when a set of options are available to it, the agent can compute a set of hypothesis based on the probable consequence of performing each option, the ultimate behavior of the agent is therefore determined as a function of its preferences and the probability assigned.} as the possessor of the ability to reason rationally.

From the cognitive psychology literature Fodor \cite{fodor76} explains cognition as computations over syntactically structured symbols or representations governed by a set of rules. This paradigm known as the language-of-thought (LoT), attempts to explain cognitive processes such as perception, language-learning, rational choice, \etc, cognitively/computationally . A limitation of such rule-based systems occurs when performing inductive learning in complex systems having only sparse and noisy examples. In this case, the set of rules can easily blow-up combinatorially, in an attempt to faithfully explain all instances encountered. There was therefore the need to handle systematically uncertainties, by use of probabilistic models.

The work by Goodman \etal\cite{goodmanGFT07} provided a unifying theory to quantitatively relate the rule-based, combinatorial LoT with a rational statistical approach to concept learning, but unlike their studies,  work presented in \cite{goodmanGFT07}, where the focus is primarily on general Boolean concept learning, we propose a very specific computational framework for a multimodal rational agent to learn inductively about the concepts of 3-D objects in a scale and viewpoint invariant manner, when presented with few and noisy visual and/or haptics sensory data.

\subsection{Multisensory conceptual modeling}
Mounting evidence from the cognitive and neuroscience literature supporting the metamodal hypothesis, suggest that \emph{core object recognition and categorization} is solved in the brain via a cascade of reflexive, largely feedforward computations in the inferior temporal cortex, which extend beyond the domain of vision to other biological senses (e.g., touch, audition, olfaction)\cite{diCarlo12}. DiCarlo \etal in \cite{diCarlo12} explored how the brain computes this recognition solution by considering the problem at different levels of abstraction. Yildrim and Jacobs \cite{yildirimJ12} explored how object recognition knowledge could be transferred across the visual and haptic modalities via a feedforward computational model and supported their claims with corresponding human experiments.

Quiroga \etal \cite{Quiroga05} reported on a subset of human medial temporal lobe (MTL) neurons that are selectively activated by strikingly different pictures of given individuals, landmarks or objects and in some cases even by letter strings with their names. As an example, a specific neuron fired selectively to pictures of the movie star Halle Berry and responded also to the letter string {\tt HALLE BERRY} (but not to other names of known and unknown people, animals and places). The selective responses of these neurons could still be triggered by stimuli in other sensory modalities, such as the name of a person pronounced by a synthesized voice\cite{quiroga12}. Quiroga thus argued that the brain contains ``concept cells'', involved in the representation of individual people or objects regardless of the modality used to sense those people or objects. Similarly, Konkle \etal \cite{Konkle09} hypothesized that the same neural region processes motion regardless of the modality through which the motion was sensed. They found that motion aftereffects transferred between vision and touch - when adapted to visual motion in a certain direction, people felt tactile motion aftereffects in the opposite direction, and vice versa. These results thus suggest the clear distinction between (i) the internal representations of the external world, (ii) the stimulus or sensation (\eg the image impinging on the retina) and (iii) their perception or the interpretations given to the stimulus.

Why would the brain possess representations and operations that are shared by multiple sensory modalities? It is possible that these representations
and operations efficiently support perception in a world where many basic parameters of sensory stimulation - space, time, and intensity - co-vary across multiple sensory signals. By developing a computational framework that takes advantage of statistical properties across modalities,  objects are treated as concepts and are characterized in terms of their intrinsic 3-D properties, rather than by only their sensory properties.

The major goal of this work therefore, is to develop a realistic model that integrates the major themes on concepts and concept learning as listed in Section \ref{sec:concepts}, by (a) identifying a set of rational rules that govern the cognitive processes of specifying classes or categories of 3-D objects; (b) transforming these rules into a probabilistic generative grammar for 3-D objects (the concept language); (d) generating a compositional hypothesis space of candidate concepts from the grammar; (e) developing a realistic multimodal perception-driven likelihood function to relate each hypothesis to sensory object data; and lastly, (f) performing Bayesian inference over the grammatically structured hypothesis space. We demonstrate the effectiveness of the model by successfully making several levels of inferences (such as categorizing objects, providing viewpoint information on objects when encountered in the wild, \etc) on abstract objects when presented with multisensory object data.

The closest work to our proposed methodology is that of Yildrim and Jacobs \cite{yildirimJ12}, where they explored how object recognition knowledge could be transferred across the visual and haptic modalities and supported their claims with corresponding human experiments. The major difference between this work and theirs is that their focus was very much on showing \emph{experimentally} in humans that object category learning does truly transfer between modalities, and then building a simplistic computational model to support their findings.

\section{Fribbles and the \emph{See \& Grasp Dataset}}\label{sec:fribbles}
\begin{figure*}[htbp]
  \begin{center}
    \begin{tabular}{cccccccc}
      \includegraphics [width=0.075\linewidth]{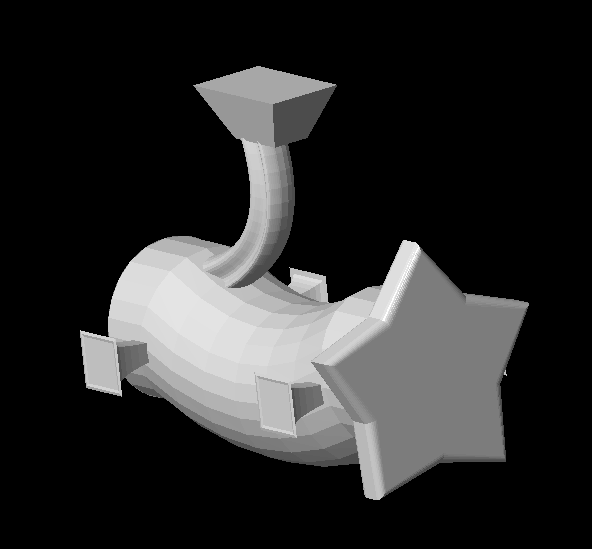} \hspace*{-1.5mm}&
      \includegraphics [width=0.075\linewidth]{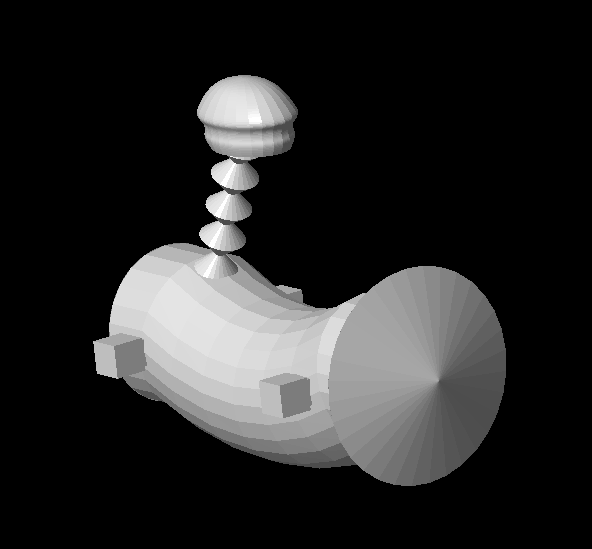} \hspace*{-1.5mm}&
      \includegraphics [width=0.075\linewidth]{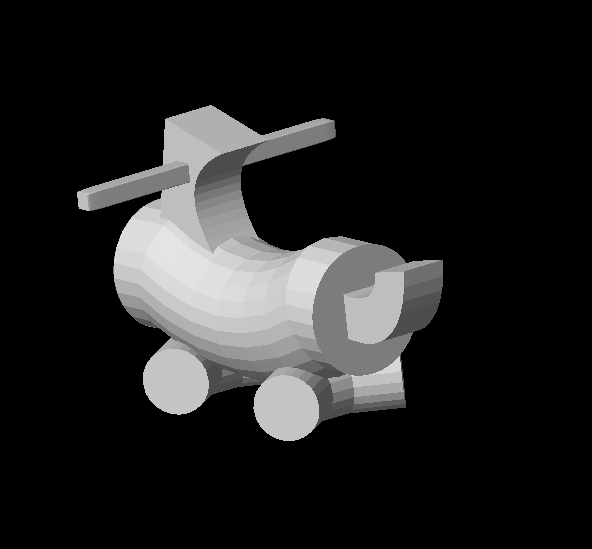} \hspace*{-1.5mm}&
      \includegraphics [width=0.075\linewidth]{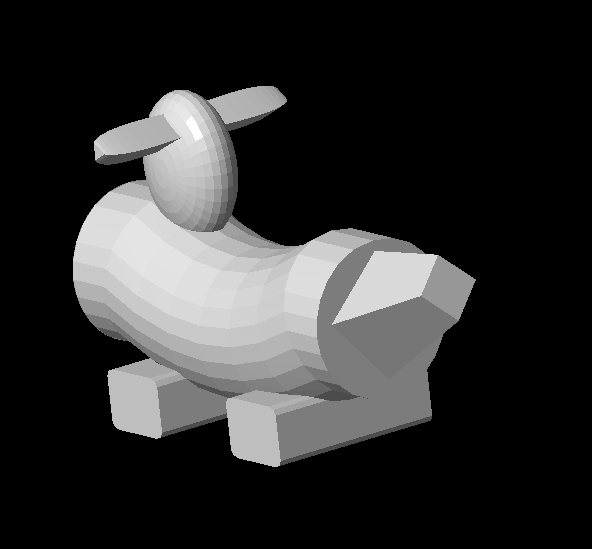} \hspace*{-1.5mm}&
      \includegraphics [width=0.075\linewidth]{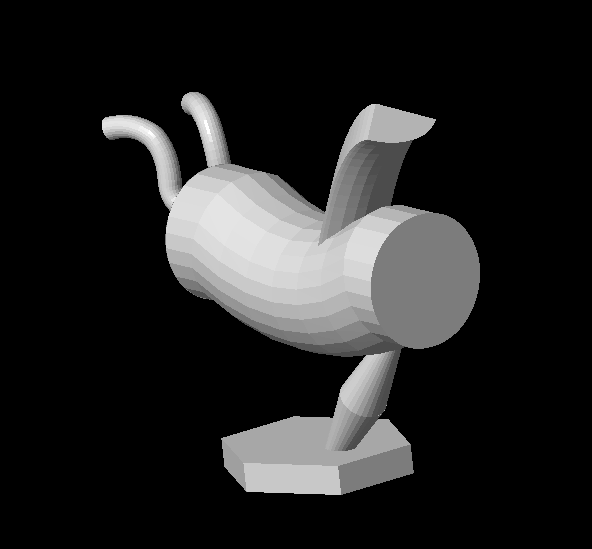} \hspace*{-1.5mm}&
      \includegraphics [width=0.075\linewidth]{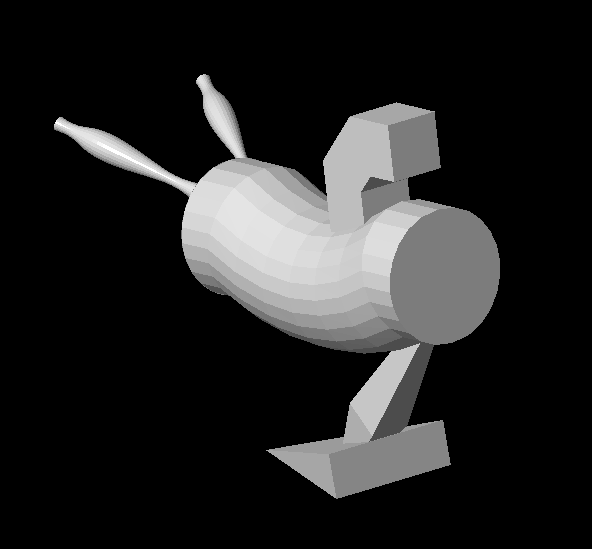} \hspace*{-1.5mm}&
      \includegraphics [width=0.075\linewidth]{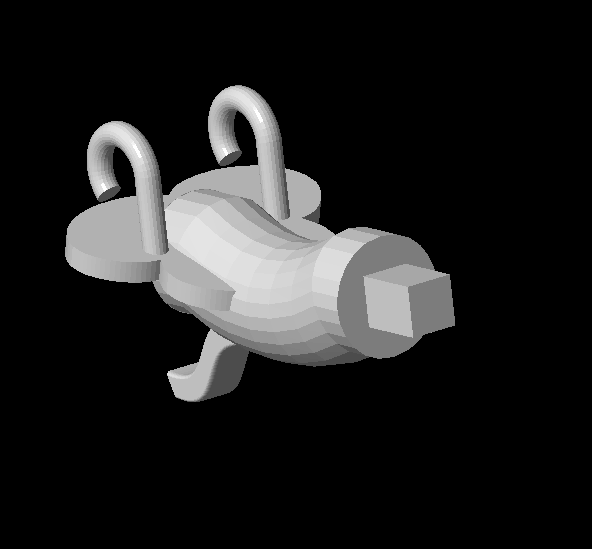} \hspace*{-1.5mm}&
      \includegraphics [width=0.075\linewidth]{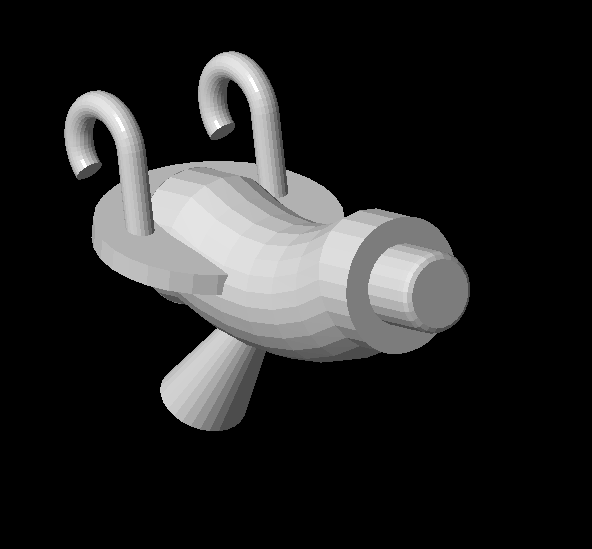} \\
      \includegraphics [width=0.05\linewidth]{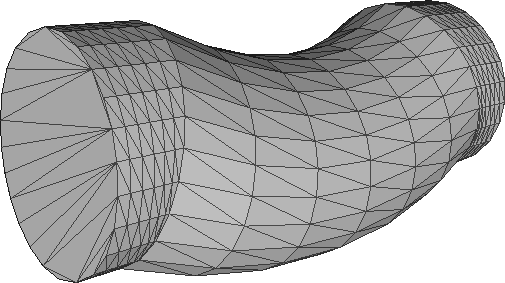} \hspace*{-1.5mm}&
      \includegraphics [width=0.05\linewidth]{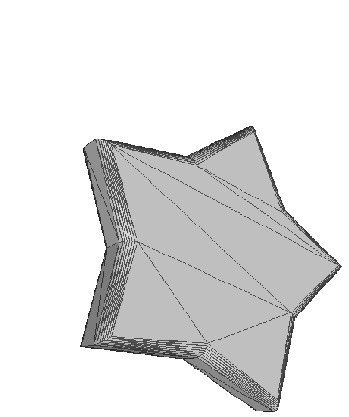} \hspace*{-1.5mm}&
      \includegraphics [width=0.05\linewidth]{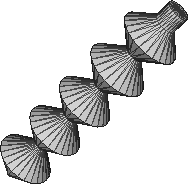} \hspace*{-1.5mm}&
      \includegraphics [width=0.05\linewidth]{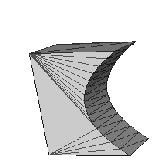} \hspace*{-1.5mm}&
      \includegraphics [width=0.05\linewidth]{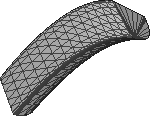} \hspace*{-1.5mm}&
      \includegraphics [width=0.05\linewidth]{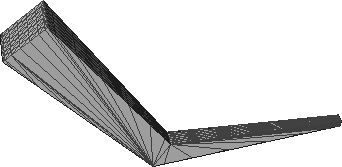} \hspace*{-1.5mm}&
      \includegraphics [width=0.05\linewidth]{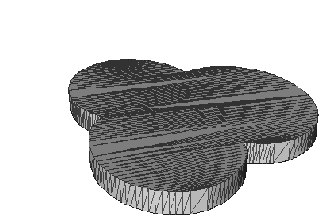} \hspace*{-1.5mm}&
      \includegraphics [width=0.05\linewidth]{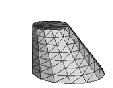}
    \end{tabular}
    \caption{The top row shows eight examples of fribbles, where pairs of columns belong to one category. The second row shows some of the constituent parts that fribbles are made of.}\label{fig:fribsamples}
  \end{center}
\end{figure*}

In our study, we assume that objects are characterized completely by their overall shape and are categorized by the different parts (and their locations) that make up the object. We perform our multisensory object recognitions study on a set of artificially generated 3-D objects known as \emph{fribbles}. Fribbles were originally developed by \cite{Tarr98visualobject} for the study of visual object recognition and the subset we are using in this study was made publicly available by \cite{yildirimJ12}. Fribbles are complex, multi-part 3-D objects with categorical structures, and as such are ecologically valid as objects that can be used in building visual-haptic computational models (such as we propose). One major advantage of using fribbles is the fact that the simulated visual and haptic renderings for our computational model are perfectly matched, given that they were 3-D printed from the same 3-D object models. The dataset consists of 40 fribbles organized into four categories with 10 exemplars per category and the category prototypes differ in their constituent parts. An exemplar for a category is created by perturbing the category prototype primarily in terms of its constituent parts. Each constituent part exists in only one spatial location so we use this feature to simplify our model in Section \ref{sec:model}. Each fribble object in the dataset is associated with a 3-D object model and an image of the fribble from a canonical viewpoint clearly showing the parts of the fribble. Figure \ref{fig:fribsamples} illustrates eight such fribbles, two from each of the four categories. A small subset of the constituent parts that make up the fribbles is also shown.

\section{Computational Model for Multisensory  Object Perception}\label{sec:model}
The goal of the computational model is to learn a scale invariant, viewpoint invariant, multisensory representation of a 3-D object using raw visual and haptics sensory data. The choice of the two sensory modalities we investigate are based on the claims (with corresponding human experiments) reported by Yildrim and Jacobs \cite{yildirimJ12}, that object recognition knowledge is successfully shared and transferred across the visual and haptic modalities.
It is important to note that that our definition of haptics here is quite limited and we only consider the joint angles of a simulated hand grasping an object as our haptic object features.

Connecting our computational model with the notion of concepts, rational rules and LoT described in Section \ref{sec:concepts}, we first define a grammar $\mathcal{G}$ to encode our beliefs about what the concept of 3-D fribble objects are comprised of. The rational rules governing the concept are therefore the production rules of the grammar, and since we have no strong biases to favor one part over another, the probabilities of the production rules at each level are uniform in our probabilistic context-free grammar (pCFG). Lastly, in order to obtain production from our concept, we derive computations over the structured symbols in $\mathcal{G}$ governed by a set of probabilistic rules, thus implementing a probabilistic LoT. Thus our concept grammar is a probabilistic process for generating 3-D fribble-like objects. That object is generated as a particular sequence of production rules which we call a derivation $\mathcal{D}$ (or a parse tree), and it specifies the constituent parts $\mathcal{P}$ that make up the object. Each derivation $D$ is therefore a hypothesis for a fribble object, proposed by the concept grammar $\mathcal{G}$. Given the observable sensory data $X$, we can now apply Bayes' rule to define a posterior distribution over the object representations as:

\begin{equation}
P(\mathcal{P}, \mathcal{D}|\mathcal{G}, X) = P(\mathcal{P},\mathcal{D}|\mathcal{G})P(X|\mathcal{D},\mathcal{P}) \label{eqn:bayes}
\end{equation}

\noindent $P(\mathcal{P},\mathcal{D}|\mathcal{G})$ represents the derivation-and-parts syntactic prior while $P(X|\mathcal{D},\mathcal{P})$ is the sensory-based likelihood distribution. In the ensuing parts of this section, we will describe our implementation of the concept grammar, and our specific derivations of the components of Equation \ref{eqn:bayes}

\subsection{Concept grammar}
We define our pCFG $\mathcal{G}$, to consist of a set of nonterminal symbols $\cal{N}$, a set of terminal symbols $\cal{T}$ and a set of probabilistic productions $\cal{R}$. In addition, we also define a set of \emph{preterminals} $\mathcal{N}_p$, as those nonterminal that can only be rewritten into a single terminal symbol. The proposed grammar defines at least one derivation structure over every terminal in the language. Furthermore, the grammar defines a probability distribution over all possible derivations. The grammar contains the start symbol $F$ and six nonterminals \{N M M1 M2 M3 M4\} out of which there are four preterminals \{M1 M2 M3 M4\}. Lastly, there are forty-seven terminal symbols, each corresponding to a constituent 3-D object part. A few examples of constituent parts are shown in the second row of Figure \ref{fig:fribsamples}. Every fribble is made up a trunk and four other parts. Interestingly, with the exception of the trunk, constituent parts are not shared across fribble categories and each part is always placed at the same location in relation to the fribble trunk. We encode much of this information in the simple grammar shown in Figure \ref{fig:grmr}

\begin{figure}[h!t]
\begin{center}
\begin{tabular}{c}
\fbox{
    \parbox{3.50in}{
\footnotesize
\smallskip
$F \longrightarrow N \: P5 $\\
$N \longrightarrow N M | N M M | N M M M | M M M M $\\
$M \longrightarrow M1 | M2 | M3 | M4 $\\
$M1 \longrightarrow P4|P12|P13|P16|P24|P25|P30|P35|P38|P46|P47 $\\  
$M2 \longrightarrow P1|P7|P9|P14|P18|P21|P27|P31|P32|P39|P44 $\\  
$M3 \longrightarrow P2|P8|P11|P17|P20|P23|P26|P29|P33|P37|P42 $\\ 
$M4 \longrightarrow P3|P6|P10|P19|P22|P15|P28|P34|P36|P40|P43|P45 $\\ 
}}
\end{tabular}
\end{center}
\vspace*{-0.15in}
\caption{An illustration of the production rules for the 3-D fribble grammar}\label{fig:grmr}
\end{figure}\vspace*{-0.2in}

\subsection{Derivation-and-parts syntactic prior}\label{sec:partsprior}
We factorize our syntactic prior so that $P(\mathcal{D},\mathcal{P}|\mathcal{G}) = P(\mathcal{D}|\mathcal{G})P(\mathcal{P}|\mathcal{D})$
Each production choice in a derivation has a probability assigned to it, so that the probability of the derivation is the product of of the probabilities of all the choices. For each nonterminal symbol expansion, the set of production probabilities $\tau$ sum to 1. The probability of a derivation $D$ is :
\begin{eqnarray}
P(\mathcal{D}|\mathcal{G}, \tau) = \prod_{n \in \mathcal{D}} \tau(n) \label{eqn:prior2}
\end{eqnarray}

\noindent where $n\in \mathcal{D}$ are the production choices that make up the derivation and $\tau(n)$ is the probability of each choice. Equation \ref{eqn:prior2} shows a strong bias towards simple derivations where longer (and possibly more accurate) derivations are penalized by having lower probabilities. Thus, $\tau$ is marginalized out and we make use of a noninformative prior (based on the principle of indifference), which assigns equal probabilities to all possibilities. We model our noninformative prior distribution over derivations using the rational rules model presented in \cite{goodmanGFT07}:
\begin{eqnarray}
P(\mathcal{D}|\mathcal{G}) = \prod_{s \in \mathcal{N}} \frac{\beta(\mathbf{C}_s(\mathcal{D})+\mathbf{1})}{\beta(\mathbf{1})} \label{eqn:prior3}
\end{eqnarray}

\noindent where $\beta(\cdot)$ is the multinominal beta function, $\mathbf{C}_s(\mathcal{D})$ is the count of how many times each production rule $s$ of a nonterminal in $\mathbf{N}$ is used in the derivation $\mathcal{D}$ and $\mathbf{1}$ is a vector of ones.

The other component of the prior probability is the parts-based prior $P(\mathcal{P}|\mathcal{D})$, which is generated from the set of terminal symbols in a derivation. Our grammar has 47 such symbols corresponding to 47 unique fribble constituent parts. The trunk is the only constituent part shared across fribble categories and in general, a constituent part exists at only one spatial location in relation to the trunk. The only exception to this is in the case of an aggregate part where the multiple instances of the same part are located at pre-specified positions with known orientations. For these reasons, each terminal $P1\ldots P47$ in $\mathcal{G}$  is realized as (i) a 3-D constituent part object file; (ii) a spatial location of the center of the part in relation to the center of the trunk; (iii) the number of times an instance of the part occurs; and (iv) the orientation of the part (or its multiple instances when required) in relation to the trunk. 
With so many degrees-of-freedom, in this model/grammar, we have chosen to encode the part, location and orientation information in the expansion of the terminal symbol to its 3-D realization.

As observed in the grammar specification (Figure \ref{fig:grmr}), it is possible for a derivation to use the same preterminal symbol over and over again, so that it is perfectly legal for multiple parts to co-exist at the same location. Therefore, to discourage our model from using each preterminal more than once, we introduce the parts-driven prior probability:
\begin{eqnarray}
P(\mathcal{P}|\mathcal{D}) = \prod_{s \in \mathcal{N}_p} \frac{1}{|\mathcal{N}_p|} = \frac{1}{{\mathcal{N}_p}^{|s|}} \label{eqn:prior4}
\end{eqnarray}

\noindent where $s \in \mathcal{N}_p$ are the preterminals in the derivation; and $|s|$ is the number of times they are used more than once and $|\mathcal{N}_p| =4$, the total number of preterminals in $\mathcal{G}$. The combined prior distribution $P(\mathcal{D},\mathcal{P}|\mathcal{G})$  favors smaller trees as well, but penalizes reuse of preterminals, which also leads to a faster convergence during simulation. But the inclusion of the parts-based component reduces the noninformative property of our overall prior.

\subsection{Sensory-based likelihood function}
Multisensory object representations do not make direct make direct contact with sensory data since they are by nature modality-independent. Hence, we combine object recognition knowledge with the visual and haptic modalities via sensory-specific feed-forward models. The vision-specific and haptic-specific models are used to calculate the likelihood function $P(X|\mathcal{D},\mathcal{P})$ from Equation \ref{eqn:bayes}. These sensory-specific models are the only way in which the perceived sensory data can affect the learning we are doing about object representations (or more formally, the posterior distribution in Equation \ref{eqn:bayes}). To compute the likelihood function we define a sensory feature term $F(\mathcal{D},\mathcal{P})$.

\subsubsection{Haptic likelihood model}
\begin{figure}
\begin{center}
\vspace*{-4mm}
\begin{tabular}{cc}
      \includegraphics[width=0.3\linewidth]{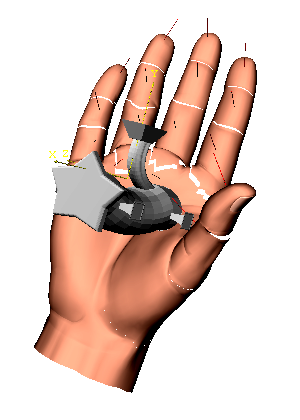} & \hspace*{-7mm}
      \includegraphics[width=0.3\linewidth]{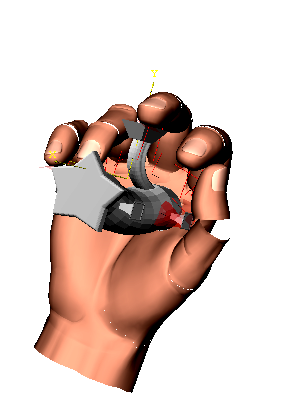} \\
\end{tabular}
\caption{GraspIt! simulation showing an open hand in the starting position and the final grasp where the hand auto-closes around a fribble object}\label{fig:graspit}
\end{center}
\end{figure}
Estimating the haptics likelihood involved the use of a simulated robot hand, GraspIt! \cite{graspit}, created at the University of Columbia\footnote{installation files at \url{http://graspit.sourceforge.net/}}.
Given the derivation from the concept grammar, the 3D object is generated as above and converted to the vrml markup language using an open source mesh converter\footnote{mesh converter code can be found at \url{http://www.cs.princeton.edu/~min/meshconv/}}. The markup file is passed to an open simulated robot hand that auto-grasps the 3D object defined in the file. Figure \ref{fig:graspit} illustrates how the simulated robot hand is in a default open-hand position and then auto-closes around the fribble object for a grasp, yielding haptic joint angle features.

In our case, the haptic-specific feature $F(\mathcal{D},\mathcal{P})$ is realized as the sixteen joint angles (robot hand degrees-of-freedom) required to fully grasp the object by GraspIt! The haptic likelihood function is calculated using the cosine-distance (\textit{dCos}) between the rendered object's haptics vector and the haptic vector from the perceived sensory data $F(X)$.  $0 \le \textit{dCos} \le 1$; when $\textit{dCos}\to 1.0$  the haptic vectors are very dissimilar and vice versa when \textit{dCor} is close to zero. We therefore compute $1-\emph{dCos}$ as the distance between $F(X)$ and $F(\mathcal{D},\mathcal{P})$ for every rotation. The maximum distance value is the likelihood value.

\newpage
\subsubsection{Vision likelihood model}
\begin{wrapfigure}{r}{2.0in}
\begin{center}
\begin{tabular}{cc}
   \includegraphics[width=0.5\linewidth]{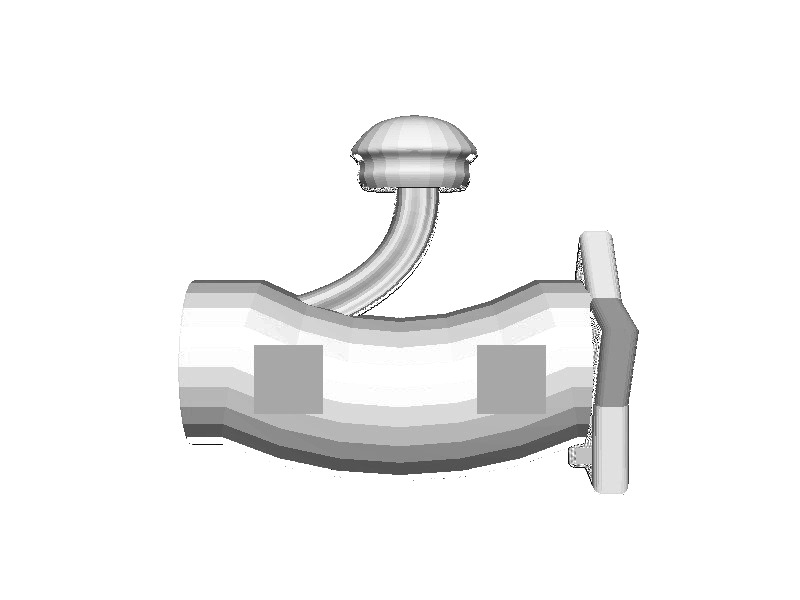} & \hspace*{-5mm}
   \includegraphics[width=0.5\linewidth]{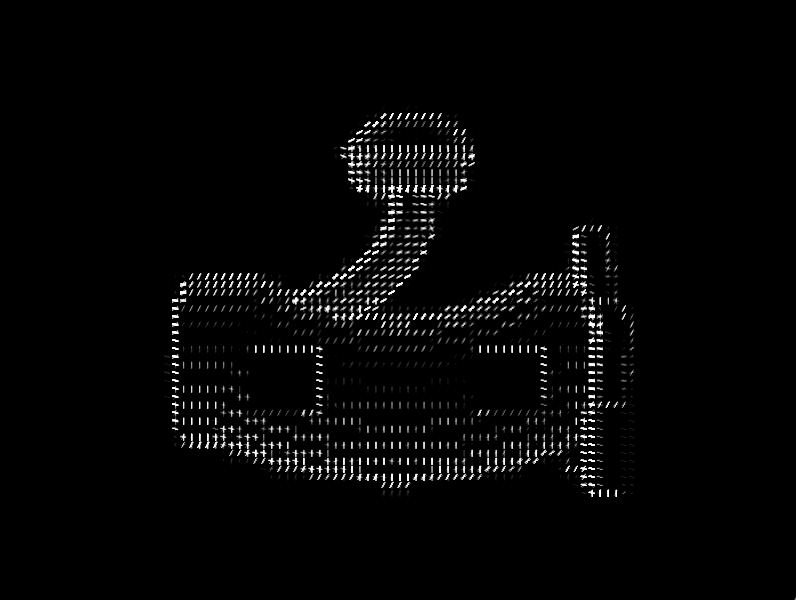}
\end{tabular}
   \caption{Left: a fribble training sample; right: HoG feature emphasizing its contour}\label{fig:hog}
\end{center}
\end{wrapfigure}

Estimating the visual likelihood involved rendering the 3-D representation realized from a specific derivation into 2D images, using Panda3D, an open source game engine written in python that includes a graphics library. The terminal symbols in a derivation are realized into a 3-D object as explained previously in Section \ref{sec:partsprior}. The 3-D object is then mapped to 2D at many different pre-specified viewpoints, by rotating 360 degrees in intervals of 40 degrees along the heading direction and from -25 through zero to +25 degrees along the pitch axis. No rotations are done along the roll axis. Using a standard camera coordinate and fixed diffused lighting in the graphics engine, we took 2D snapshots of the fribbles at each new rotation.

To realize $F(\mathcal{D},\mathcal{P})$ for the vision-specific model, we implemented a contour-based, multi-resolution representation known in the computer vision literature as the Histogram-of-Gradients (HoG) \cite{HOG05}\footnote{HoG is a technique where the occurrences of gradient orientation are counted in localized portions of an image, and is computed on a dense grid of uniformly spaced cells, with overlapping local contrast normalization for improved accuracy}. To define the likelihood function, we compute a correlation distance (\emph{dCor}) between the HoG features of the perceived sensory data $F(X)$ and the sensory feature obtained from the 2D rendering $F(\mathcal{D},\mathcal{P})$, after applying basic image processing routines to roughly align the two. $0 \le \textit{dCor} \le 1$; when $\textit{dCor}\to 1.0$  the feature vectors are very similar and vice versa when \textit{dCor} is close to zero. We compute \emph{dCor} between $F(X)$ and the features obtained for every rotated images and select the maximum value as the vision-based likelihood value.

\subsection{Tree-based MCMC Metroplolis-Hastings algorithm}
Performing exact inference on Equation \ref{eqn:bayes} is intractable, so we rely on Markov chain Monte Carlo methods for sampling from the posterior distribution. The goal here is to define a Markov chain in the space of grammar derivations (or parse trees) and we employ the Metropolis Hastings (M-H) algorithm to accomplish this using sub-tree generation \cite{goodmanGFT07}. In this approach, current derivations are modified to generate new ones, by randomly selecting a nonterminal node in the derivation $\mathcal{D}$, removing all the nodes below it and regenerating a new proposal derivation $\mathcal{D}'$ from that node, using the probabilistic rules of the grammar. The proposal is accepted with the probability :
\begin{equation}
A = \min\left\{1, \frac{P(X|\mathcal{D}',\mathcal{P}')}{P(X|\mathcal{D},\mathcal{P})} \cdot \frac{P(\mathcal{D}'|\mathcal{G})}{P(\mathcal{D}|\mathcal{G})} \cdot \frac{|\mathcal{N}_{p_\mathcal{D}|}}{|\mathcal{N}_{p_\mathcal{D'}}|} \right\}\label{eqn:accept}
\end{equation}
\noindent where $|N_{p_D}|$ is the number of nonterminals in the derivation $D$. This algorithm is a valid metropolis-Hastings sampler whose stationary distribution is the posterior distribution over the multisensory representations in our multimodal model.

\vspace*{-0.15in}
\section{Experiments and Results}
There are several inference questions that we are interested in addressing with the learned generative multimodal model, and these include:\\
\noindent (1) Given the learned distribution over the space of fribbles, what do the ``fantasy'' or prototype fribbles look like?
(2) Given the grasp of a fribble that has never been seen by the model,  can the model predict the category of the fribble?
(3) Given the 2-D image of a fribble that has ever been presented to the model, and at an unknown orientation and scale, can the model predict the category of the fribble?

In order to investigate the questions above, in our experiments, we trained the multimodal model with 24 out of the total of 40 fribbles, 6 from each of 4 categories. All the original fribble objects from the \emph{See \& Grasp dataset} were scaled down by 0.3 for all our analysis. For the simulations, each MCMC chain was run for 10,000 iterations and the first 1,000 samples were discarded as burn-in. Each iteration computed sensory likelihood values for 27 rotation angles (9 in the heading direction and 3 in the pitch direction; roll was not considered).

Once the model was trained, samples were drawn and evaluated qualitatively. Next, for new object classification, we presented the model with grasp joint angles from the remaining 16 fribbles, \ie the grasps of 4 novel fribbles -never been seen by the model- from each of 4 categories.

\begin{figure*}[htbp]
  \begin{center}
    \begin{tabular}{cccccccc}
      \includegraphics [width=0.1\linewidth]{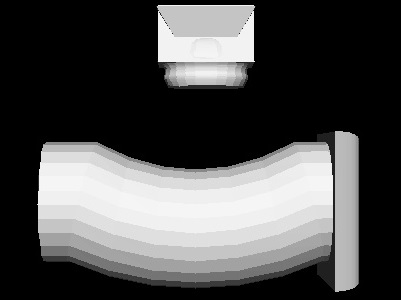} \hspace*{-4.5mm}&
      \includegraphics [width=0.1\linewidth]{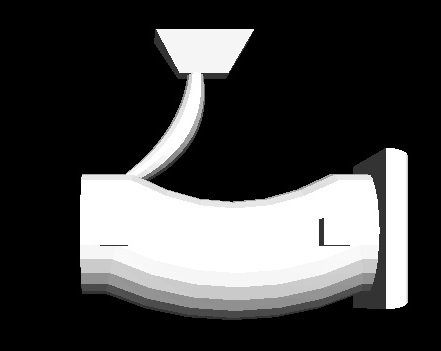} &
      \includegraphics [width=0.1\linewidth]{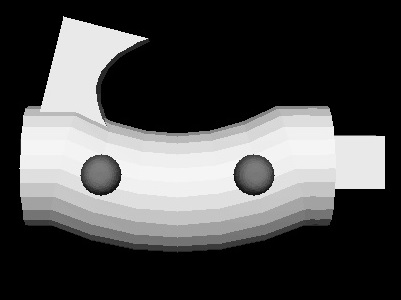} \hspace*{-4.5mm}&
      \includegraphics [width=0.1\linewidth]{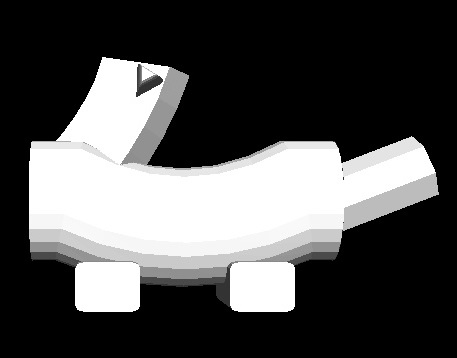} &
      \includegraphics [width=0.1\linewidth]{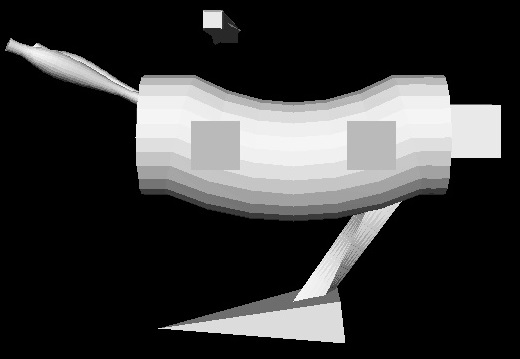} \hspace*{-4.5mm}&
      \includegraphics [width=0.1\linewidth]{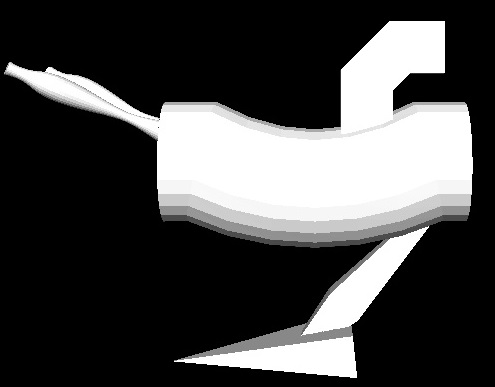} &
      \includegraphics [width=0.1\linewidth]{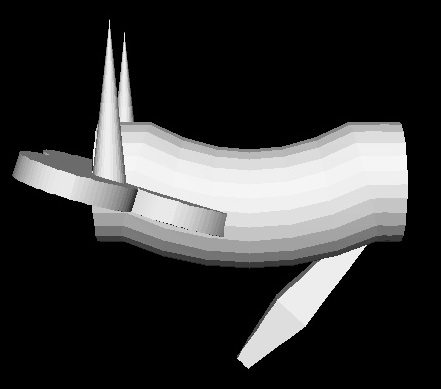} \hspace*{-4.5mm}&
      \includegraphics [width=0.1\linewidth]{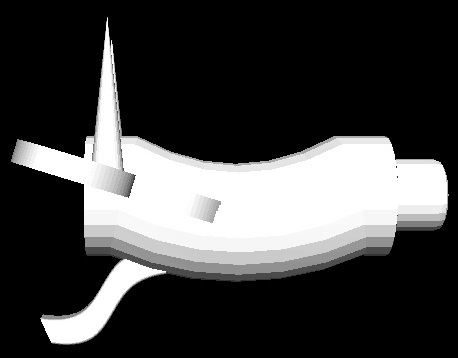}
    \end{tabular}
    \caption{Pairs of images showing ``fantasy fribble'' (left of the pair) and the closest resembling actual sample (right of the pair) for each of the four fribble categories. These ``fantasy fribbles'' are the learned model prototypes.}\label{fig:prototypes}
  \end{center}
\end{figure*}

\noindent\textbf{Sampling from the learned model}\\
The prototypical shape for each category of objects is the 3-D shape with the largest posterior probability. In general, for all the categories, the model learned 3-D, part-based representations that bear strong resemblance to example training objects. Figure \ref{fig:prototypes} shows pairs of images where the sample drawn from the model is on the left and the closest representative training example is shown on the right. Interestingly, each prototype has one or more missing parts, thus suggesting the need to better balance the effects of the prior and the likelihood functions in our model. The model as it currently stands appears to give more significance to the prior over the likelihood, thus displaying a strong preference for smaller derivations as observed across the prototypes.

\subsection{Multimodal fribble categorization}
\subsubsection*{Vision-based categorization}
\textbf{Place results of vision based categorization here....}
Select each of the 16 test samples
For each test sample, perturb the orientation and scale of the test fribble
get the prototype of each of the 4 classes
Generate the 27 images at different orientations and use the HOG to compute which
class this image is closest to
Set that class value as the class of this test fribble.
Complete test and enter table here....

\subsubsection*{Haptics-based categorization}
Figure \ref{fig:confmatrix} shows the categorization results using both haptics information. We select the four test examples from each category and present each one to GraspIt! in order to generate the haptic sensory data (synonymous with the agent touching the objects). The resulting grasp joint angles are presented to the model, and the sensory grasp joint angles are compared to the grasp joint angles of each of the 4 model prototypes. The test grasp is therefore classified as belonging to the category of the prototype that it is closest in distance to. The resulting confusion matrix is shown in Figure \ref{fig:confmatrix} .
\begin{figure}
    \begin{center}
    \vspace*{-4mm}
    \begin{tabular}{r c c c c}
     & \textbf{C1} & \textbf{C2} & \textbf{C3 } & \textbf{C4}  \\
    \cline{2-5}
    \textbf{C1} &\multicolumn{1}{|c|}4 &\multicolumn{1}{|c|} 0 &\multicolumn{1}{|c|} 0 &\multicolumn{1}{|c|} 0 \\
    \cline{2-5}
    \textbf{C2} &\multicolumn{1}{|c|}0 &\multicolumn{1}{|c|} 3 &\multicolumn{1}{|c|} 1 &\multicolumn{1}{|c|} 0 \\
    \cline{2-5}
    \textbf{C3} &\multicolumn{1}{|c|}0 &\multicolumn{1}{|c|} 2 &\multicolumn{1}{|c|} 2 &\multicolumn{1}{|c|} 0 \\
    \cline{2-5}
    \textbf{C4} &\multicolumn{1}{|c|}0 &\multicolumn{1}{|c|} 0 &\multicolumn{1}{|c|} 0 &\multicolumn{1}{|c|} 4  \\
    \cline{2-5}
    \end{tabular}
    \caption{Confusion matrix showing the haptics-based categorization results on test objects, approximately 81\%  accuracy (13 of 16 correctly categorized) }\label{fig:confmatrix}
    \end{center}
\end{figure}
\normalsize

\section{Discussion and Future Work}
In this study, we successfully developed a multimodal learning framework that integrated major themes on concepts and concepts learning by using a probabilistic generative grammar on which we performed inference via Bayesian induction. We performed our multisensory object recognitions study on a set of artificially generated 3-D objects, fribbles, and successfully performed scale-invariant, pose-invariant object categorization by comparing test samples with the 4-class prototypes (sampled 3D objects with the largest posterior probability). When clean, known samples are presented to the model, it gives an accuracy of approximately 71\% with haptic sensory data.

In the future, rather than only comparing the test cases to the sample prototypes (thus assuming that the test case was generated from the model), it will be useful to directly present the test cases to the model to determine the nature of parse trees that will be generated from the sensory data. This way the likelihood of an object being a fribble or not can also be computed. Lastly, it will be interesting to perform an in-depth study of how scale changes affect haptics, and at what scale the grasp simulator will fail to distinguish between object categories. On a grander scale, can we now develop computational models for \emph{``any notion''} once we can encode its rational rules and can compute likelihood measures?

\newpage
{\normalfont
\bibliographystyle{aaai}
\nocite{*}
\bibliography{aaai14}
}

\end{document}